\documentclass[11pt]{article}

\usepackage[preprint]{acl}

\usepackage{times}
\usepackage{latexsym}
\usepackage{times}
\usepackage{latexsym}
\usepackage{amsmath}
\usepackage{booktabs}
\usepackage{multirow}
\usepackage{multicol}
\usepackage{amssymb}
\usepackage[T1]{fontenc}
\usepackage[utf8]{inputenc}
\usepackage{microtype}
\usepackage{inconsolata}
\usepackage{graphicx}
\usepackage{enumitem}

\usepackage[table]{xcolor}
\definecolor{mygo}{RGB}{197, 224, 180}
\newcommand{\cc}{\cellcolor{mygo}} 

\usepackage{siunitx}    

\usepackage[T1]{fontenc}

\usepackage[utf8]{inputenc}

\usepackage{microtype}

\usepackage{inconsolata}

\usepackage{graphicx}
\usepackage{amsmath,amssymb}
\title{VIB-Probe: Detecting and Mitigating Hallucinations in Vision-Language Models via Variational Information Bottleneck}

\author{First Author \\
  Affiliation / Address line 1 \\
  Affiliation / Address line 2 \\
  Affiliation / Address line 3 \\
  \texttt{email@domain} \\\And
  Second Author \\
  Affiliation / Address line 1 \\
  Affiliation / Address line 2 \\
  Affiliation / Address line 3 \\
  \texttt{email@domain} \\}

\begin{document}

\author{
    \textbf{Feiran Zhang} \thanks{Equal contribution.}\textbf{,} \
    \textbf{Yixin Wu} \footnotemark[1]\textbf{,} \
    \textbf{Zhenghua Wang,} \
    \textbf{Xiaohua Wang,} \\
    \textbf{Changze Lv,} \
    \textbf{Xuanjing Huang}\textbf{,} \
    \textbf{Xiaoqing Zheng}\thanks{Corresponding Author.} \\
    College of Computer Science and Artificial Intelligence, Fudan University, Shanghai, China \\
    Shanghai Key Laboratory of Intelligent Information Processing \\
    {\tt\small $\{$frzhang25,yixinwu23$\}$@m.fudan.edu.cn}
    {\tt\small $\{$xjhuang,zhengxq$\}$@fudan.edu.cn}
}

\maketitle



\begin{abstract}
Vision-Language Models (VLMs) have demonstrated remarkable progress in multimodal tasks, but remain susceptible to hallucinations, where generated text deviates from the underlying visual content. 
Existing hallucination detection methods primarily rely on output logits or external verification tools, often overlooking their internal mechanisms. 
In this work, we investigate the outputs of internal attention heads, postulating that specific heads carry the primary signals for truthful generation.
However, directly probing these high-dimensional states is challenging due to the entanglement of visual-linguistic syntax and noise. 
To address this, we propose VIB-Probe, a novel hallucination detection and mitigation framework leveraging the Variational Information Bottleneck (VIB) theory. 
Our method extracts discriminative patterns across layers and heads while filtering out semantic nuisances through the information bottleneck principle. 
Furthermore, by leveraging the gradients of our VIB probe, we identify attention heads with strong causal influence on hallucinations and introduce an inference-time intervention strategy for hallucination mitigation. 
Extensive experiments across diverse benchmarks demonstrate that VIB-Probe significantly outperforms existing baselines in both settings. Our code will be made publicly available.
\end{abstract}

\section{Introduction}


Vision-Language Models (VLMs) have emerged as an influential force in multimodal artificial intelligence, demonstrating a sophisticated ability to generate contextually rich natural language descriptions grounded in visual patterns \cite{DBLP:journals/corr/abs-2304-14178,DBLP:journals/corr/abs-2502-13923,DBLP:conf/icml/0008LSH23}. By integrating visual encoders with large language models, VLMs have shown impressive performance across diverse vision-language tasks, including image captioning, visual question answering, and multimodal machine translation \cite{DBLP:journals/corr/abs-2304-08485,DBLP:conf/iclr/Zhu0SLE24,https://doi.org/10.48550/arxiv.2306.15195,DBLP:journals/corr/abs-2402-08360}. Despite these advancements, VLMs remains prone to hallucinations, where generated descriptions are unfaithful to the objects or relations present in the source image \cite{DBLP:journals/chinaf/YinFZXWSSLSC24,DBLP:conf/acl/HeZGFHJ0CW25}. This lack of visual fidelity undermines the reliability and applicability of VLMs, particually in high-stakes domains that demand precise multimodal reasoning and factual accuracy.

\begin{figure}[t!]
    \hspace*{0.8cm}
    \includegraphics[width=0.87\linewidth]{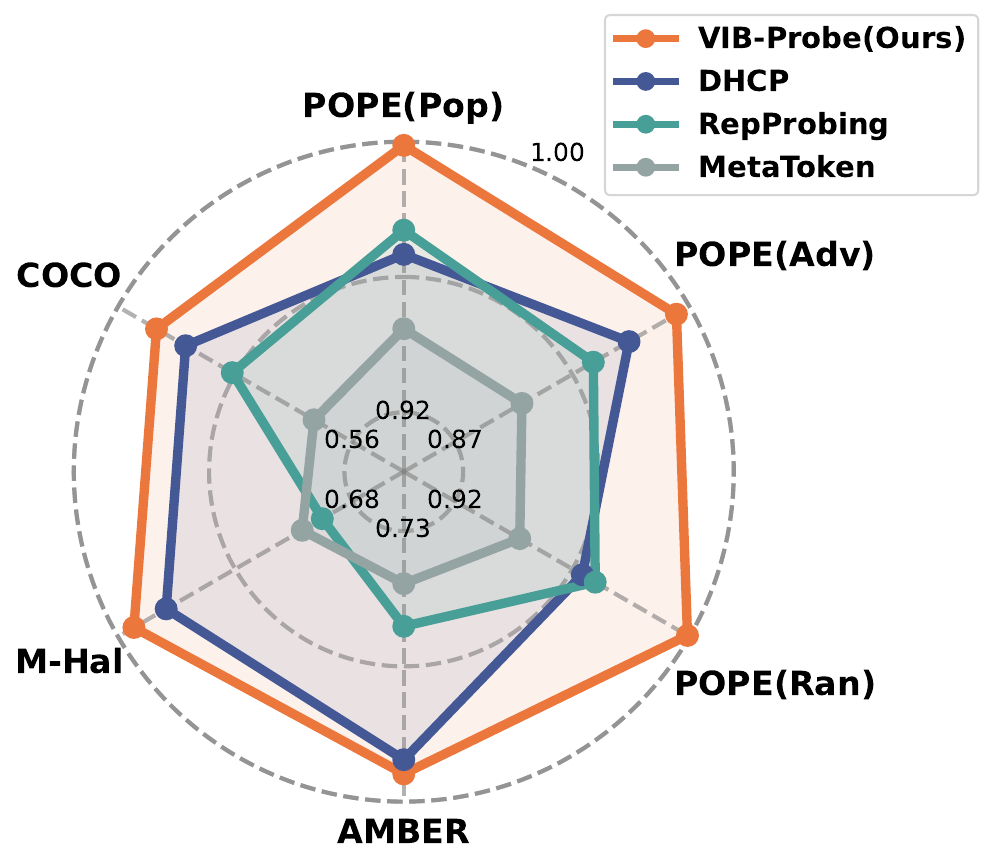}
    \vspace{-2mm}
    \caption{Hallucination detection performance comparison across 6 benchmarks, based on the AUPRC metric. Our proposed VIB-Probe consistently achieves state-of-the-art overall results.}
    \vspace{-3mm}
    \label{fig:radar}
\end{figure}

Existing approaches to hallucination detection primarily rely on surface-level confidence indicators, such as logit-based entropy or divergence \cite{Fieback_2025,https://doi.org/10.48550/arxiv.2504.12137,https://doi.org/10.48550/arxiv.2505.11741,DBLP:conf/iclr/HendrycksG17}. These heuristic-based classifiers typically exploit only a narrow slice of the model’s internal dynamics and depend on manually engineered features that may fail to generalize across diverse architectures. 
Consequently, developing robust and efficient mechanisms for detecting hallucinations in VLM outputs remains a significant open challenge.

Recent research in interpretability suggests that VLM hallucinations are often rooted in fragile attention dynamics introduced by the visual modality \cite{DBLP:conf/cvpr/JiangCZLSY25,DBLP:conf/cvpr/TangLXHHX0PYZLL25,DBLP:conf/iclr/JiangKPG25,DBLP:conf/iclr/YangL0X25}. Specifically, a model may attend to irrelevant regions, infer non-existent objects, or over-rely on linguistic priors at the expense of visual grounding \cite{DBLP:journals/corr/abs-2510-20229}. 
Crucially, this informational drift is not confined to the final output layer, but rather emerges progressively across internal layers \cite{Zhang_2025,DBLP:conf/acl/HeZGFHJ0CW25}. Hallucination-related signals are encoded within the outputs of specific attention heads across layers, while these signals are often entangled with task-irrelevant syntactic noise.

\begin{figure*}[t]
    \centering
    \includegraphics[width=1\textwidth]{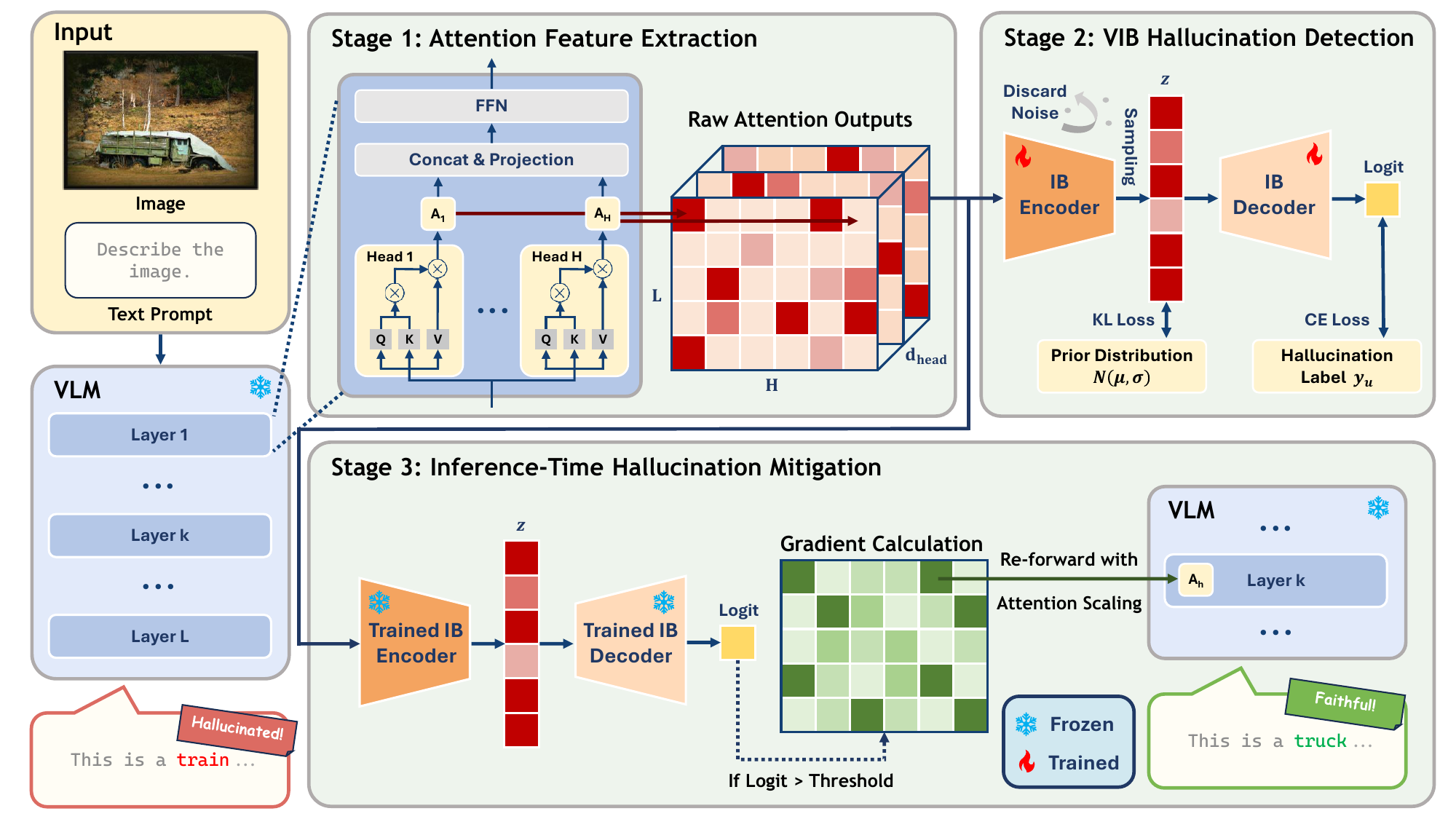}
    \caption{Overview of the VIB-Probe framework. The Information Bottleneck (IB) theory is leveraged to detect and mitigate hallucinations by probing internal attention features. \textbf{Stage 1:} We extract raw output vectors from all attention heads across all Transformer layers ($L \times H$) during VLM decoding. \textbf{Stage 2:} The extracted features are fed into an IB Encoder, which compresses the high-dimensional input into a compact latent representation $z$. This process filters out task-irrelevant noise while retaining minimal sufficient statistics for prediction. \textbf{Stage 3:} Leveraging the trained VIB modules, inference-time mitigation is achieved by suppressing the attention heads with high risks of hallucination for each token, producing a more faithful output generation.}
    \label{fig:main}
\end{figure*}

Motivated by these insights, we propose \textbf{VIB-Probe}, a novel framework grounded in \textbf{Variational Information Bottleneck (VIB)} theory \citep{DBLP:journals/corr/physics-0004057,DBLP:conf/iclr/AlemiFD017}. 
As illustrated in Figure~\ref{fig:main}, VIB-Probe distills a compact latent representation from the high-dimensional attention head outputs across all Transformer layers, retaining information predictive of hallucinations while suppressing noise and spurious correlations. We employ a multi-layer encoder to capture the bottleneck features for robust detection. 
Furthermore, we extend our approach to hallucination mitigation by applying gradient-based attribution from the probe's logits to the attention heads. By these means, we identify specific ``hallucination-sensitive'' heads that exert strong causal influence on unfaithful generation. We then introduce an inference-time mitigation strategy that selectively suppresses these heads during decoding when the detected hallucination risk exceeds a predefined threshold. Extensive experiments across multiple benchmarks demonstrate that our approach yields consistent gains in both detection and mitigation across diverse VLM architectures.

The contributions of this study can be summarized as follows:
\begin{itemize}
    \item We introduce VIB-Probe, a novel framework for hallucination detection that exploits the information of multi-layer, multi-head attention outputs in VLMs. By grounding our approach in Variational Information Bottleneck theory, we distill a compact yet highly predictive latent representation, enabling robust detection across both open-ended generation and closed-form QA settings.
    \item We propose a training-free, inference-time mitigation strategy that bridges the gap between detection and control. By employing probe-based attribution, we identify hallucination-sensitive attention heads and dynamically suppress their outputs upon high risks of hallucination.
    \item We conduct comprehensive experiments across both discriminative and generative hallucinatory benchmarks, demonstrating that VIB-Probe achieves state-of-the-art performance in hallucination detection and mitigation, while further highlighting its robustness and generalizability across diverse perturbations and architectures.
\end{itemize}

\section{Related Work}

\subsection{Hallucinations in VLMs}
Vision-Language Models (VLMs) integrate visual encoders with Large Language Models (LLMs) via projection layers to enable multimodal reasoning \cite{DBLP:journals/corr/abs-2304-08485}. 
Compared to factual errors in text-only LLMs, VLM hallucinations mainly arise from failures in visual grounding. 
They are commonly grouped into \textbf{object}, \textbf{attribute}, and \textbf{relational} hallucinations \cite{DBLP:conf/iclr/ZhouCYZDFBY24}.

\paragraph{Hallucination Detection}
Early detectors relied on shallow output statistics (e.g., token confidence or entropy), which often generalize poorly under complex reasoning. 
Reference-free methods aim to verify outputs without external evidence \citep{Li_2024, prabhakaran-etal-2025-vade}. 
Recent work probes mechanistic signals in attention, e.g., Lookback Lens~\citep{chuang2024lookbacklensdetectingmitigating} and OPERA~\citep{Huang_2024}, by analyzing aggregated attention patterns during decoding, improving discrimination between grounded and hallucinatory outputs. 
Building on this direction, we move beyond raw attention weights and apply VIB to attention head outputs to better isolate hallucination-relevant signals from high-dimensional noise.

\paragraph{Hallucination Mitigation}
Mitigation methods are typically categorized into \textbf{training-based}, \textbf{post-generation}, and \textbf{inference-time} approaches. 
Training-based methods enhance robustness via instruction-tuning on curated data \citep{DBLP:journals/corr/abs-2304-08485,DBLP:conf/eccv/ZhangWZLZ24,DBLP:conf/iclr/ZhouCYZDFBY24}, while post-generation methods employ external verifiers for iterative refinement \citep{DBLP:journals/chinaf/YinFZXWSSLSC24}. 
Inference-time interventions have gained attention for avoiding retraining costs: VCD~\citep{DBLP:conf/cvpr/LengZCLLMB24} reduces reliance on linguistic priors via visual perturbation, and PAI~\citep{DBLP:conf/eccv/LiuZC24} and IBD~\citep{DBLP:conf/cvpr/ZhuJCXY025} strengthen visual grounding by adjusting attention to image tokens. 
Our method follows this paradigm but introduces gradient-based attribution to target hallucination-sensitive heads for training-free intervention.

\subsection{Information Bottleneck Theory}
The Information Bottleneck (IB) principle~\citep{DBLP:journals/corr/physics-0004057} serves as a robust information-theoretic framework for regularizing internal representations. 
By compressing model input to minimize mutual information, IB encourages the model to discard irrelevant features while retaining essential semantic content, thereby enhancing generalization capabilities. 
This principle has been extensively adopted across various machine learning paradigms, 
including image generation~\citep{DBLP:journals/corr/abs-2510-20165}, generative classification~\citep{DBLP:conf/nips/ArdizzoneMRK20}, explanation regeneration~\citep{DBLP:conf/acl/Li0KB23}, and retrieval-augmented generation~\citep{Zhu_2024}. 
To operationalize the IB objective in deep neural networks, \citet{DBLP:conf/iclr/AlemiFD017} introduced the Variational Information Bottleneck (VIB). 
Inspired by the architecture of Variational Autoencoders (VAEs)~\citep{DBLP:journals/corr/KingmaW13}, VIB employs a variational approach to approximate the IB trade-off and has demonstrated significant efficacy in parsing~\citep{Li_2019}, low-resource fine-tuning~\citep{DBLP:conf/iclr/MahabadiBH21}, and graph structure learning~\citep{DBLP:conf/aaai/Sun0P0FJY22} domains.

\section{Method}

\subsection{Preliminaries}

\paragraph{Attention Head Output}
For vision-language models of the most prevalent LLaVA-style \cite{DBLP:journals/corr/abs-2304-08485} architecture, a vision encoder is coupled with a decoder-only large language model via a projection layer. An input image is encoded into a sequence of visual tokens, which are projected into the LLM's embedding space and concatenated with the textual prompt tokens. This multimodal sequence is processed by $L$ Transformer decoder layers, each containing $H$ attention heads. During each autoregressive decoding step $t$, the model predicts the next token conditioned on the image, the prompt, and the previously generated tokens.
For a given layer $l \in [1, L]$ and head $h \in [1, H]$, the input hidden states $\mathbf{X}^l$ are transformed into query, key, and value matrices:
\begin{equation}
{Q}_{l,h} = {X}^l {W}_{l,h}^Q, {K}_{l,h} = {X}^l {W}_{l,h}^K, {V}_{l,h} = {X}^l {W}_{l,h}^V
\label{eq:qkv}
\end{equation}
where ${W}^Q, {W}^K, {W}^V \in \mathbb{R}^{d_{model} \times d_{h}}$ are the projection weights.
The attention weights $\mathbf{A}_{l,h}$ are then computed via the scaled dot-product:
\begin{equation}
\mathbf{A}_{l,h} = \text{softmax}\left(\frac{\mathbf{Q}_{l,h} (\mathbf{K}_{l,h})^\top}{\sqrt{d_{h}}}\right)
\label{eq:attention_weight}
\end{equation}

To capture the raw, disentangled information flow prior to the final head-mixing, we extract the \textbf{pre-projection attention head output} $\mathbf{O}_{l,h}$:
\begin{equation}
\mathbf{O}_{l,h} = \mathbf{A}_{l,h} \mathbf{V}_{l,h}
\label{eq:attention_output}
\end{equation}

For each token generated at step $t$, we aggregate $\mathbf{O}_{l,h}$ across all layers and heads to construct a representation tensor $\mathcal{T} \in \mathbb{R}^{L \times H \times d_{h}}$. This tensor provides a comprehensive ``snapshot'' of the model's internal multimodal processing and serves as the primary input for our VIB-Probe framework.

\paragraph{Information Bottleneck}
The Information Bottleneck principle defines an optimal representation $\mathbf{z}$ of an input signal $\mathbf{v}$ that maximizes its predictive power regarding a target $\mathbf{y}$ while minimizing the information retained from $\mathbf{v}$. 
Formally, the IB objective is formulated as the following constrained optimization:
\begin{equation}
\min \mathcal{L}_{\mathrm{IB}}
= \beta\, I(\mathbf{v};\mathbf{z}) - I(\mathbf{z};\mathbf{y}),
\label{eq:ib}
\end{equation}
where $I(\cdot;\cdot)$ denotes mutual information and $\beta>0$ is a Lagrange multiplier controlling the trade-off between \emph{compression} (minimizing $I(\mathbf{v};\mathbf{z})$) and \emph{prediction} (maximizing $I(\mathbf{z};\mathbf{y})$).
By penalizing $I(\mathbf{v};\mathbf{z})$, the model is forced to discard ``semantic nuisances'' features that are irrelevant to the grounding of visual content.

Directly optimizing Eq.~\eqref{eq:ib} is generally intractable, as computing mutual information requires knowledge of the underlying data distributions.
The \textbf{Variational Information Bottleneck} addresses this by introducing a variational upper bound on the compression term $I(\mathbf{v};\mathbf{z})$ and replaces the predictive term with a tractable likelihood model.
Specifically, VIB parameterizes an encoder $p_{\theta}(\mathbf{z}\mid \mathbf{v})$ and a decoder $p_{\phi}(\mathbf{y}\mid \mathbf{z})$, and uses a prior $r(\mathbf{z})$.
The resulting objective is:
\begin{equation}
\begin{split}
\min \mathcal{L}_{\mathrm{VIB}}
= &\beta\, \mathbb{E}_{\mathbf{v}}\!\left[\mathrm{KL}\!\left(p_{\theta}(\mathbf{z}\mid \mathbf{v}) \,\|\, r(\mathbf{z})\right)\right] \\
&+ \mathbb{E}_{\mathbf{v}}\, \mathbb{E}_{\mathbf{z}\sim p_{\theta}}\!\left[-\log p_{\phi}(\mathbf{y}\mid \mathbf{z})\right],
\label{eq:vib}
\end{split}
\end{equation}
where the first term acts as a \emph{compression} regularizer and the second term represents the negative log-likelihood loss of \emph{prediction}. 
In practice, for binary labels, the prediction loss is implemented as a binary cross-entropy (BCE) loss.

\subsection{Hallucination Detection via VIB on Attention Head Outputs}
\label{sec:detector}
Building on prior observations, we propose VIB-Probe, a lightweight detector based on Information Bottleneck theory. VIB-Probe is designed to aggregate the internal holistic attention information of VLMs for hallucination detection.

\paragraph{Problem Setup}
Given an input image and a text prompt, a VLM generates a total of $N$ tokens auto-regressively. 
At each decoding step $u$, we extract the pre-projection attention head outputs from all layers and heads (Eq.~\eqref{eq:attention_output}), stacking them into a tensor $\mathcal{T} \in \mathbb{R}^{L\times H\times d_h}$. 
Our goal is to predict a binary hallucination label $\mathbf{y}_u \in \{0,1\}$, where $\mathbf{y}_u=1$ denotes a hallucination and $\mathbf{y}_u=0$ signifies those visually-grounded.
The resulting training set is defined as $\mathcal{D}=\{(\mathcal{T}_u,\mathbf{y}_u)\}_{u=1}^N$.

\paragraph{VIB Detector Architecture}
We treat the tensor $\mathcal{T}$ as the raw internal signal $\mathbf{v}_u$ and feed it into a lightweight convolutional or multi-layer perceptron encoder $f_{\psi}(\cdot)$ to extract a high-level feature representation $\mathbf{h}_u\in\mathbb{R}^{d_f}$:
\begin{equation}
\mathbf{v}_u := \mathcal{T},\qquad
\mathbf{h}_u = f_{\psi}(\mathbf{v}_u),
\label{eq:cnn_feat}
\end{equation}

A variational bottleneck then parameterizes an approximate posterior $q_{\psi}(\mathbf{z}_u\mid \mathbf{v}_u)$ as a multivariate diagonal Gaussian:
\begin{equation}
\begin{split}
q_{\psi}(\mathbf{z}_u\mid \mathbf{v}_u)
&= \mathcal{N}\!\left(\boldsymbol{\mu}_{u}, \mathrm{diag}(\boldsymbol{\sigma}^2_{u})\right),
\quad \\
[\boldsymbol{\mu}_{u}, \log \boldsymbol{\sigma}^2_{u}] &= g_{\psi}(\mathbf{h}_u),
\label{eq:posterior}
\end{split}
\end{equation}
where the decoder $g_{\psi}$ is a single or multiple fully-connected layers.
Similar to \cite{Li_Eisner_2019}, we sample $\mathbf{z}_u$ during training using the reparameterization trick \cite{DBLP:journals/corr/KingmaW13} to ensure end-to-end differentiability:
\begin{equation}
\mathbf{z}_u = \boldsymbol{\mu}_{u} + \boldsymbol{\sigma}_{u} \odot \boldsymbol{\epsilon},
\quad \boldsymbol{\epsilon}\sim \mathcal{N}(\mathbf{0},\mathbf{I}).
\label{eq:reparam_det}
\end{equation}

At inference time, we adopt a deterministic approach for stability, utilizing the mean representation $\mathbf{z}_u = \boldsymbol{\mu}_{u}$ for prediction.
Finally, a linear classification layer computes the \textbf{hallucination risk logit} $s_u$ and the corresponding probability $\hat{p}_u$ through the sigmoid function $\sigma(\cdot)$:
\begin{equation}
s_u = w^\top \mathbf{z}_u + b,\qquad
\hat{p}_u = \sigma(s_u),
\label{eq:pred_prob}
\end{equation}

\paragraph{Training Objective}
Following the Variational Information Bottleneck principle, we optimize the latent representation $\mathbf{z}_u$ to be maximally informative about the label $\mathbf{y}_u$ while remaining minimally sufficient with respect to the input $\mathbf{v}_u$. 
We regularize the information flow by penalizing the KL divergence between the approximate posterior $q_{\psi}(\mathbf{z}_u\mid \mathbf{v}_u)$ and a standard normal prior $r(\mathbf{z})=\mathcal{N}(\mathbf{0},\mathbf{I})$. The token-level detection loss is formulated as:

\begin{equation}
\begin{split}
\mathcal{L}_{\mathrm{det}}
= &\mathbb{E}_{(\mathbf{v}_u,\mathbf{y}_u)\sim\mathcal{D}}
\Big[
\underbrace{\mathrm{BCE}(\mathbf{y}_u,\hat{p}_u)}_{\text{prediction}}\\
&+
\beta\underbrace{\mathrm{KL}\!\left(q_{\psi}(\mathbf{z}_u\mid \mathbf{v}_u)\,\|\,r(\mathbf{z})\right)}_{\text{compression}}
\Big],
\label{eq:det_loss}
\end{split}
\end{equation}
where $\beta>0$ is a Lagrange multiplier that controls the trade-off between prediction accuracy and representation compression. The first term here is the standard binary cross-entropy (BCE) loss:
\begin{equation}
\mathrm{BCE}(\mathbf{y}_u,\hat{p}_u)
= -\mathbf{y}_u\log \hat{p}_u -(1-\mathbf{y}_u)\log (1-\hat{p}_u),
\label{eq:bce}
\end{equation}

Given our choice of a diagonal Gaussian posterior (Eq.~\eqref{eq:posterior}), the KL term has a closed form:

\begin{equation}
\begin{split}
\mathrm{KL}\!\left(\mathcal{N}(\boldsymbol{\mu}_{u}, \mathrm{diag}(\boldsymbol{\sigma}^2_{u})) \,\|\, \mathcal{N}(\mathbf{0},\mathbf{I})\right)
=\\
\frac{1}{2}\sum_{i=1}^{d_z}\left(\mu_{u,i}^2 + \sigma_{u,i}^2 - \log \sigma_{u,i}^2 - 1\right),
\label{eq:kl_det}
\end{split}
\end{equation}
where $d_z$ denotes the dimension of the bottleneck latent space. 
During training, we minimize the objective function $L_{det}$ with respect to the parameters of the encoder $f_{\psi}$ and decoder $g_{\psi}$. At inference time, the raw logit $s_u$ is utilized to assess hallucination risk and further mitigation.

\subsection{Hallucination Mitigation}

Building upon the trained VIB detector, we propose an inference-time mitigation strategy that translates detection signals into actionable model control. By attributing the predicted hallucination risk to specific internal components, we can dynamically suppress the most influential attention heads that leads to hallucinations. 

At each decoding step $u$, we perform a VLM forward pass to extract the attention head outputs $\mathcal{T} \in \mathbb{R}^{L\times H\times d_h}$ and compute the VIB hallucination risk logit $s_u$. If $s_u\le\tau$ (where $\tau$ is a risk threshold), the model samples the next token normally. If $s_u > \tau$, an intervention is triggered to rectify the potential hallucination, by modifying attention heads and regenerating the token.

\paragraph{Gradient-based Attribution and Head Selection}

To identify which heads contribute most to hallucination risks, we perform a backward pass through the \textbf{frozen} VIB detector. We compute the gradient of the risk logit by each attention head at the current step: 
$g^{l,h} = \nabla_{o^{l,h}} s_u$.
Since our intervention involves scaling the head outputs by a coefficient $\alpha^{l,h}$, such that the modified output becomes 
$\tilde o^{l,h}=\alpha^{l,h}o^{l,h}$. The sensitivity of the risk logit to this scaling is:
\begin{equation}
\nabla_{\alpha^{l,h}} s_u = \left\langle g^{l,h},\,o^{l,h}\right\rangle.
\end{equation}

We define the \textbf{head importance score} as the magnitude of this sensitivity: $I^{l,h}=|\langle g^{l,h},o^{l,h}\rangle|$. 
We then select the set of most influential heads $\mathcal{K}=\mathrm{TopK}(\{I^{l,h}\})$ for targeted suppression.

\paragraph{Inference-Time Single-Step Head Suppression}
We initialize all the output scaling coefficients as $\alpha^{l,h}=1$. For the heads identified in $\mathcal{K}$, we apply a single-step suppression update to reduce hallucinatory risk:
\begin{equation}
\alpha^{l,h}\leftarrow 1-\lambda\cdot \mathrm{ReLU}\!\left(\left\langle g^{l,h},\,o^{l,h}\right\rangle\right),
\quad (l,h)\in\mathcal{K},
\end{equation}
where $\lambda$ is a hyperparameter for controlling the suppression strength. We keep $\alpha_t^{l,h}=1$ unmodified for $(l,h)\notin\mathcal{K}$.
Finally, we rerun the VLM decoding step using the modified head outputs $\tilde o^{l,h}=\alpha^{l,h}o^{l,h}$ to obtain the edited logits and then sample the regenerated token. 

\begin{table*}[t]
    \small
    \centering
    \tabcolsep=1.15mm

\begin{tabular}{llcccccccccc}
\toprule
\multirow{2}{*}{\textbf{Benchmark}} 
& \multirow{2}{*}{\textbf{Method}} 
& \multicolumn{2}{c}{\textbf{MiniGPT-4}} 
& \multicolumn{2}{c}{\textbf{LLaVA-v1.5}}
& \multicolumn{2}{c}{\textbf{LLaVA-v1.6}}
& \multicolumn{2}{c}{\textbf{Qwen2.5-VL}}
& \multicolumn{2}{c}{\textbf{Average}} \\
\cmidrule(lr){3-4} \cmidrule(lr){5-6} \cmidrule(lr){7-8} \cmidrule(lr){9-10} \cmidrule(lr){11-12}
& 
& \textbf{A-ROC} & \textbf{A-PR} 
& \textbf{A-ROC} & \textbf{A-PR} 
& \textbf{A-ROC} & \textbf{A-PR}
& \textbf{A-ROC} & \textbf{A-PR} 
& \textbf{A-ROC} & \textbf{A-PR} \\

\midrule
\rowcolor{gray!15}
\multicolumn{12}{l}{\textit{\textbf{Discriminative Benchmarks}}} \\
\midrule

\multirow{6}{*}{POPE}
& AvgEnt  
& 76.27 & 68.64 & 77.43 & 67.52 & 79.51 & 70.66 & 78.99 & 70.20 & 78.05 & 69.26 \\
& AvgProb
& 61.56 & 63.39 & 64.25 & 63.90 & 63.06 & 66.44 & 68.28 & 64.55 & 64.29 & 64.57 \\
& RepProbing 
& 91.18 & 92.30 & 94.68 & 94.50 & 93.01 & 93.87 & 96.82 & 95.89 & 93.92 & 94.14 \\
& MetaToken 
& 89.69 & 90.07 & 93.07 & 92.22 & 94.21 & 94.33 & 94.10 & 94.84 & 92.77 & 92.87 \\
& DHCP 
& 93.80 & 91.76 & 94.87 & 94.53 & 94.92 & 94.20 & 96.80 & \textbf{96.52} & 95.10 & 94.25 \\
& \cc \textbf{VIB-Probe (ours)}
& \cc \textbf{94.19} & \cc \textbf{93.37} & \cc \textbf{96.52} & \cc \textbf{96.96} & \cc \textbf{95.99} & \cc\textbf{95.51}
& \cc \textbf{96.98} & \cc 96.40 & \cc \textbf{95.92} & \cc \textbf{95.56} \\
\midrule

\multirow{6}{*}{AMBER}
& AvgEnt 
& 61.25 & 58.53 & 62.05 & 62.43 & 65.48 & 62.20 & 66.42 & 66.80 & 63.80 & 62.49 \\
& AvgProb
& 59.68 & 55.90 & 64.74 & 60.28 & 64.81 & 63.79 & 64.51 & 63.33 & 63.44 & 60.83 \\
& RepProbing 
& 72.25 & 71.11 & 77.53 & 76.82 & 75.35 & 74.74 & 74.61 & 74.52 & 74.94 & 74.30 \\
& MetaToken 
& 74.18 & 73.39 & 74.60 & 75.20 & 74.46 & 74.81 & 75.59 & 75.10 & 74.71 & 74.63 \\
& DHCP 
& 83.18 & 82.27 & 82.07 & 81.89 & 84.77 & 83.64 & 84.77 & 83.98 & 83.70 & 82.95 \\
& \cc \textbf{VIB-Probe (ours)} 
& \cc \textbf{83.40} & \cc \textbf{82.94} & \cc \textbf{82.95} & \cc \textbf{82.43} & \cc \textbf{85.99} & \cc \textbf{85.91}
& \cc \textbf{85.51} & \cc \textbf{85.82} & \cc \textbf{84.46} & \cc \textbf{84.28} \\

\midrule
\rowcolor{gray!15}
\multicolumn{12}{l}{\textit{\textbf{Generative Benchmarks}}} \\
\midrule

\multirow{6}{*}{M-HalDetect}
& AvgEnt  
& 54.90 & 38.22 & 53.27 & 36.87 & 55.90 & 37.52 & 63.52 & 41.39 & 56.90 & 38.50 \\
& AvgProb
& 54.00 & 38.93 & 59.01 & 39.54 & 60.21 & 40.36 & 66.47 & 42.71 & 59.92 & 40.39 \\
& RepProbing 
& 78.21 & 70.04 & 77.18 & 69.80 & 77.38 & 71.20 & 80.92 & 71.13 & 78.42 & 70.54 \\
& MetaToken 
& 77.28 & 69.13 & 82.02 & 71.14 & 81.23 & 73.56 & 75.19 & 69.30 & 78.93 & 70.78 \\
& DHCP 
& 79.58 & 74.62 & 88.13 & 80.20 & 86.51 & 78.87 & 84.82 & 80.40 & 84.76 & 78.52 \\
& \cc \textbf{VIB-Probe (ours)} 
& \cc \textbf{83.33} & \cc \textbf{77.26} & \cc \textbf{89.98} & \cc \textbf{82.35} & \cc \textbf{88.36} & \cc \textbf{81.23} & \cc \textbf{85.17} & \cc \textbf{80.79} & \cc \textbf{86.71} & \cc \textbf{80.41} \\
\midrule

\multirow{6}{*}{COCO-Caption}
& AvgEnt  
& 52.08 & 30.81 & 58.93 & 32.01 & 55.89 & 34.72 & 60.21 & 35.26 & 56.78 & 33.20 \\
& AvgProb
& 55.36 & 32.67 & 54.45 & 33.92 & 58.88 & 36.95 & 59.05 & 34.18 & 56.94 & 34.43 \\
& RepProbing 
& 65.96 & 56.88 & 72.33 & 62.56 & 71.92 & 64.99 & 77.11 & 66.14 & 71.83 & 62.64 \\
& MetaToken 
& 65.70 & 55.34 & 67.28 & 58.30 & 67.23 & 59.35 & 70.89 & 61.20 & 67.78 & 58.55 \\
& DHCP 
& 69.52 & 58.13 & 74.06 & 64.99 & 74.20 & 68.17 & 74.14 & 67.64 & 72.98 & 64.73 \\
& \cc \textbf{VIB-Probe (ours)} 
& \cc \textbf{72.55} & \cc \textbf{62.82} & \cc \textbf{75.24} & \cc \textbf{66.51} & \cc \textbf{75.16} & \cc \textbf{69.32} & \cc \textbf{76.83} & \cc \textbf{70.52} & \cc \textbf{74.95} & \cc \textbf{67.29} \\

\bottomrule
\end{tabular}

    \caption{Results of hallucination detection across multiple baselines on discriminative and generative benchmarks. We report AUROC (A-ROC) and AUPRC (A-PR) as metrics and compare our method with baselines across four base VLMs (MiniGPT-4, LLaVA-v1.5-7B, LLaVA-v1.6-Mistral-7B, and Qwen2.5-VL-7B-Instruct).}
    \label{tab:detection_main_dis}
\end{table*}

\section{Experiments}

\subsection{Benchmarks}

We evaluate VIB-Probe across a diverse suite of hallucination detection benchmarks covering both discriminative and generative datasets.

\paragraph{POPE}
POPE \cite{li2023evaluating} is a standard diagnostic for VLM object hallucinations. 
For each image, the dataset provides three positive questions regarding existing objects and three negative questions. The negative samples are selected based on random sampling (\emph{Random}), global frequency (\emph{Popular}), or co-occurences with present objects (\emph{Adversarial}). 
Throughout our experiments, we utilize the official POPE dataset, which comprises a total of 9,000 questions across 1,500 images. 

\paragraph{AMBER}
AMBER \cite{wang2023amber} extends the scope of evaluation beyond POPE's objects to include \emph{attribute} and \emph{relation} hallucinations. The origin dataset contains 14,216 discriminative queries. We randomly sampled 5,000 queries from the original dataset for the experiments.

\paragraph{M-HalDetect}
M-HalDetect \cite{gunjal2024detecting} provides a more granular assessment of hallucinations in detailed responses. Based on the MS COCO \cite{lin2014microsoft} 2014 validation set, it includes 12,800 training and 3,200 validation samples. 
Responses are segmented and expert-annotated into four categories: \emph{Accurate}, \emph{Inaccurate}, \emph{Analysis}, and \emph{Unsure}. Approximately 25\% of segments are labeled as hallucinatory, presenting a challenge for fine-grained description tasks.

\paragraph{COCO-Caption}
To evaluate generative hallucinations in open-ended captioning, We randomly sampled 2,000 images from the MS COCO 2014 validation set, splitting them into training and validation subsets by an 80:20 ratio. We identify hallucinations from the image captions generated.

\subsection{Hallucination Detection}

\subsubsection{Experimental Setup}

\paragraph{Base Models and Datasets}
We evaluate the efficacy of VIB-Probe on four representative VLMs: MiniGPT-4~\citep{zhuminigpt}, LLaVA-v1.5-7B~\citep{DBLP:journals/corr/abs-2304-08485}, LLaVA-v1.6-Mistral-7B~\citep{liu2024improved}, and Qwen2.5-VL-7B-Instruct~\citep{DBLP:journals/corr/abs-2502-13923}. 
Experiments cover two extensively adopted discriminative benchmarks, POPE and AMBER (subset averages reported), alongside two generative datasets M-HalDetect and COCO-Caption. 
To assess detection performance, we report AUPRC and AUROC \cite{Davis_Goadrich_2006}. 
Detailed configurations are provided in Appendix~\ref{setting}.

\paragraph{Baselines}
We compare our method with classic methods based on model uncertainty and probing classifiers, as well as two strong baselines. 
MetaToken \cite{fieback126metatoken} trains a lightweight classifier by ensembling multiple statistical features derived from object token generation.
Meanwhile, DHCP \cite{zhang2025dhcp} detects hallucinations by training a lightweight prober that leverages cross-modal attention patterns during decoding. 
Implementation details are included in Appendix~\ref{baseline}.



\paragraph{Implementation Details}
VIB-Probe is implemented as a multi-layer MLP encoder for dimensionality reduction, followed by a simple linear decoder. The latent distribution is constrained by a standard Gaussian prior $\mathcal{N}(0, I)$. We set the bottleneck dimension $d=256$. We optimize the framework using AdamW with a learning rate of $2 \times 10^{-5}$ and a linear warm-up for the KL-divergence coefficient $\beta$, capped at $3 \times 10^{-4}$. For discriminative tasks, we extract representations from the last answer token; for generative tasks, we utilize the internal states corresponding to the final token of each sentence or annotated span.

\subsubsection{Result Analysis}
Table~\ref{tab:detection_main_dis} presents the hallucination detection performance of baselines and our VIB-Probe across both discriminative and generative benchmarks. 
Our VIB-Probe consistently outperforms existing state-of-the-art methods across the four evaluated VLMs. While achieving competitive results on the discriminative benchmarks (\textit{$+1.20\%$}), our method also demonstrates a pronounced advantage on the challenging generative tasks (\textit{$+2.84\%$}). This underscores its superior capability in detecting hallucinations within complex, free-form text. 

Among the baselines, uncertainty-based heuristics like AvgEnt and AvgProb perform reasonably on closed-set tasks but falter in generative settings. Conversely, RepProbing significantly outperforms these metrics, confirming that hidden states serve as effective indicators of visual fidelity. While MetaToken excels at object-level detection, its performance degrades on generative benchmarks, likely because its heuristic features are too specialized for object tokens to capture span-level or sentence-level relational errors. DHCP emerges as the strongest baseline, validating the utility of attention-based hallucination detection. 

\begin{figure}[t]
    \centering
    \includegraphics[width=1\linewidth]{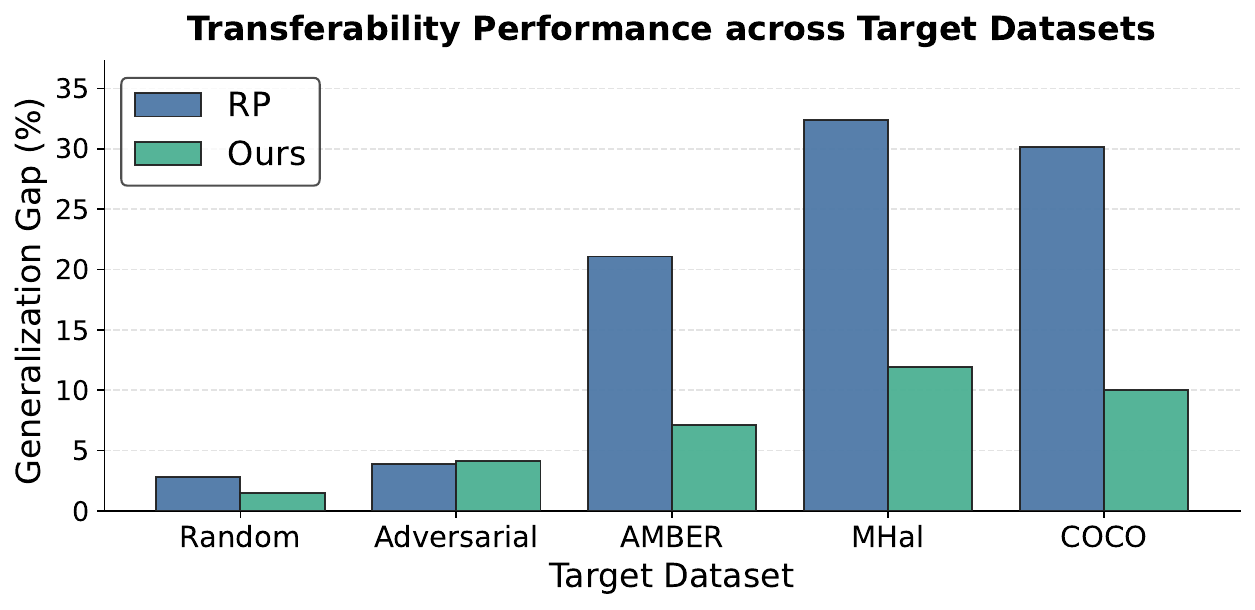}
    \caption{Generalization gap from POPE-Popular to other test sets. A lower generalization gap indicates stronger transferability performance. Results are compared based on LLaVA-v1.5-7B.}
    \label{fig:trans}
\end{figure}

\paragraph{Transferability Performance}
To evaluate the ability of VIB-Probe to extract representations highly-correlated with hallucinations that remain invariant to shifts in data distribution and task format, we conducted a series of cross-distribution and cross-task generalization experiments. 
We first assessed cross-distribution generalization by training on the POPE-Popular subset and evaluating it across all discriminative benchmarks. 
Subsequently, we evaluated cross-task generalization by evaluating the POPE-Popular detector directly on generative tasks.
As illustrated in Figure~\ref{fig:trans}, while baseline methods like RepProbing experiences significant performance degradation under domain shift (e.g., a $32.4\%$ decline on M-HalDetect), our VIB-Probe exhibits stability and stronger transferability. This indicates that the Information Bottleneck successfully distills domain-invariant hallucination signals from the internal attention dynamics, effectively filtering out dataset-specific biases.


\begin{table}[t]
\centering
\small

\begin{tabular}{l SSSS}
\toprule
\textbf{Method} 
& \multicolumn{2}{c}{\textbf{POPE}} 
& \multicolumn{2}{c}{\textbf{COCO}}  \\ 
\cmidrule(lr){2-3} \cmidrule(lr){4-5}
& \textbf{A-ROC} 
& \textbf{A-PR} 
& \textbf{A-ROC}
& \textbf{A-PR} \\ 
\midrule
AvgEnt          & 53.14 & 52.73 & 50.98 & 31.77 \\
RepProbing   & 76.92 & 77.46 & 63.23 & 54.10 \\
DHCP            & 84.77 & 83.63 & 66.40 & 58.58 \\
\cc \textbf{VIB-Probe}       & \cc \textbf{88.78} & \cc \textbf{87.30} & \cc \textbf{73.76} & \cc \textbf{64.81} \\
\bottomrule
\end{tabular}

\caption{
Robustness performance of hallucination detection on input images with random perturbations. Methods are compared based on LLaVA-v1.5-7B.
}
\label{tab:rob}
\end{table}

\paragraph{Robustness Performance}
To verify that VIB-Probe isolates compact representations specifically aligned with hallucination signals rather than low-level visual noise, we further designed a robustness experiment to evaluate its performance under varying image quality conditions. 
Specifically, we introduced random perturbations to the input images from the POPE and COCO-Caption datasets for evaluation only. These perturbations include rotation, Gaussian blur, and brightness adjustments, while ensuring that the ground-truth labels remained valid. 
Results in Table~\ref{tab:rob} demonstrate that VIB-Probe maintains high detection accuracy despite these image perturbations. This resilience indicates that our framework effectively extracts the core internal states associated with unfaithful generation, even when the model's representations are subjected to external visual noise.





\begin{table}[t]
    \small
    \centering
    \tabcolsep=1.0mm

\begin{tabular}{llcccc}
    \toprule
    \multirow{2}{*}{\textbf{Base Model}} 
    & \multirow{2}{*}{\textbf{Method}} 
    & \multicolumn{2}{c}{\textbf{POPE}}
    & \multicolumn{2}{c}{\textbf{COCO}}\\
\cmidrule(lr){3-4} \cmidrule(lr){5-6} & 
   & \textbf{ACC $\uparrow$} & \textbf{F1 $\uparrow$} 
   & \textbf{C$_i$ $\downarrow$} & \textbf{C$_s$ $\downarrow$} \\
\midrule

\multirow{5}{*}{LLaVA-v1.5-7B}
    & Vanilla     & 82.6 & 83.3 & 18.2 & 59.3 \\
    & BeamSearch  & 82.2 & 84.1 & 19.5 & 60.6 \\
    & PAI         & \textbf{84.0} & 84.6 & 14.4 & 46.7 \\
    & VCD         & 83.6 & 83.9 & 15.8 & 52.2 \\
    & \cc \textbf{VIB-Probe}   & \cc 83.7 & \cc \textbf{85.2} & \cc \textbf{14.1} & \cc \textbf{44.9} \\
\midrule

\multirow{5}{*}{LLaVA-v1.6-7B}
    & Vanilla     & 84.1 & 85.1 & 11.8 & 40.7 \\
    & BeamSearch  & 84.3 & 85.6 & 10.9 & 39.2 \\
    & PAI         & 87.9 & 88.4 & 9.2  & 35.3 \\
    & VCD         & 86.3 & 87.8 & 9.0  & 36.4 \\
    & \cc \textbf{VIB-Probe}   & \cc \textbf{88.2} & \cc \textbf{89.5} & \cc \textbf{8.7}  & \cc \textbf{32.1} \\
    \bottomrule
\end{tabular}

    \caption{Performance of hallucination mitigation on the validation sets of POPE and COCO. Methods are compared based on LLaVA-v1.5-7B.}
    \label{tab:mitigation}
\end{table}

\subsection{Hallucination Mitigation}

To validate our mitigation capabilities, we performed experiments on the POPE benchmark and a randomly selected 500-image subset of COCO val 2014. For generative evaluation, we utilized the CHAIR \cite{Rohrbach_Hendricks_Burns_Darrell_Saenko_2018} metric, which quantifies object-level hallucinations by cross-referencing generated entities against ground-truth object lists. For POPE, we reported the Accuracy and F1 score metrics. 
Experimental results in Table~\ref{tab:mitigation} indicate that while contrastive decoding-based VCD \cite{leng2024mitigating} provide a viable baseline for hallucination mitigation, inference-time attention intervention strategies such as PAI \cite{liu2024paying} generally delivers stronger performance. VIB-Probe attains the best performance across most metrics as compared to baselines, demonstrating the effectiveness of intervention on hallucination-related attention heads.

\subsection{Ablation Studies}
\paragraph{Information Bottleneck Constraint}
To verify the effectiveness of the Information Bottleneck constraint, we test a variant that retains the VIB-Probe encoder-decoder structure but removes the KL loss, solely optimizing the BCE loss. Experimental results in Table~\ref{ablation_kl} indicate that removing the KL loss degrades performance to a level comparable to the RepProbing baseline. This further demonstrates that explicitly introducing the Information Bottleneck KL divergence constraint is crucial to our gains, making our approach more effective than a simple probing classifier.

\begin{table}[t]
\centering
\small
\setlength{\tabcolsep}{1.5mm}
\begin{tabular}{l l c c}
\toprule
\multirow{1}{*}{\textbf{Base Model}} & \multirow{1}{*}{\textbf{Setting}} 
& \textbf{POPE} & \textbf{M-Hal}  \\
\midrule

\multirow{2}{*}{LLaVA-v1.5-7B}
& VIB-Probe 
& \textbf{96.96} & \textbf{82.35} \\
& $-$ KL Loss
& 88.32 & 71.91  \\

\midrule

\multirow{2}{*}{Qwen2.5-VL-7B-Instruct}
& VIB-Probe  
& \textbf{96.40} & \textbf{80.79}  \\
& $-$ KL Loss
& 92.11 & 67.34  \\

\bottomrule
\end{tabular}
\caption{Impact of removing the Information Bottleneck constraint (KL loss) on detection performance. The AUPRC metric is reported.}
\label{ablation_kl}
\end{table}

\begin{table}[t]
\centering
\small
\setlength{\tabcolsep}{3.0mm}
\begin{tabular}{llSS}
\toprule
\textbf{Base Model} & \textbf{Layers} & \textbf{POPE} & \textbf{M-Hal}  \\
\midrule

\multirow{6}{*}{LLaVA-v1.5-7B}
& \textit{All} & \textbf{96.96} & \textbf{82.35} \\
& \textit{1--8} & 68.71 & 49.66 \\
& \textit{1--16} & 73.80 & 52.39  \\
& \textit{9--24} & 91.45 & 69.18 \\
& \textit{17--32} & 93.22 & 65.94  \\
& \textit{25--32} & 89.68 & 59.44  \\

\bottomrule
\end{tabular}
\caption{Impact of layers selected for the extraction of attention head outputs on detection performance. The AUPRC metric is reported.}
\label{ablation_layer}
\end{table}

\paragraph{Layer Feature Selection}
We evaluate the impact of extracting features from a specific layers to train the VIB, rather than utilizing attention heads from all VLM layers. For LLaVA-v1.5-7B with 32 layers, results on POPE and M-HalDetect are presented in Table~\ref{ablation_layer}.
Using information from only a small subset of layers results in performance degradation, particularly on the more challenging M-HalDetect. Notably, employing only deeper layers yields better performance than using shallower layers, likely due to the fact that cross-modal information is not yet fully fused in shallow layers.




\section{Conclusion}
Hallucinations remain a formidable challenge for the deployment of Vision-Language Models in reliability-critical environments. Unfaithful generations often emerge progressively from internal attention dynamics, rather than solely from the final output. To address this, we introduce VIB-Probe, a framework that leverages high-dimensional multi-head attention outputs across all layers. By grounding our approach in the Variational Information Bottleneck theory, we effectively distill a compact latent representation that isolates hallucination-related signals from task-irrelevant noise. Beyond detection, we further demonstrate that VIB-Probe supports lightweight inference-time mitigation by identifying and down-weighting a small set of hallucination-sensitive heads upon high risks. Extensive experiments across diverse architectures and benchmarks demonstrate state-of-the-art performance in detection and mitigation, highlighting the robustness and practicality of our framework.


\section*{Limitations}
Our study primarily focuses on transformer-based vision–language models with standard attention mechanisms. 
While these architectures cover most widely used VLMs, the applicability of VIB-Probe to alternative multimodal architectures or models that do not rely on explicit attention structures has not been explored and remains an interesting direction for future work.
In addition, our method requires access to the model’s internal representations and attention outputs, which restricts it to a white-box setting and may be a potential limitation.


\bibliography{acl}

\appendix
\section{Models and Baselines}

\subsection{Vision Language Models}
\paragraph{MiniGPT-4}
MiniGPT-4\ cite{zhuminigpt} connects visual and textual modalities using a single linear projection layer. It utilizes a frozen BLIP-2 \cite{li2023blip} visual encoder, which consists of ViT-G/14 (EVA-CLIP) and a Q-Former. The language backbone is Vicuna-7B (based on LLaMA-1), comprising 32 transformer layers and 32 attention heads.

\paragraph{LLaVA-v1.5-7B}
LLaVA-v1.5 \cite{DBLP:journals/corr/abs-2304-08485} employs a two-layer MLP projector to align visual features with the language model. Its visual encoder is CLIP-ViT-L-336px. The language backbone is Vicuna-7B-v1.5 (based on Llama-2), which contains 32 layers and 32 attention heads.

\paragraph{LLaVA-v1.6-Mistral-7B}
LLaVA-v1.6 (LLaVA-NeXT) \cite{liu2024improved} introduces an "AnyRes" technique that splits high-resolution images into grids to overcome resolution limits, while still using the CLIP-ViT-L-336px visual encoder. The backbone is Mistral-7B-Instruct-v0.2, featuring 32 layers and 32 attention heads.

\paragraph{Qwen2.5-VL-7B-Instruct}
Qwen2.5-VL \cite{DBLP:journals/corr/abs-2502-13923} utilizes Naive Dynamic Resolution and M-RoPE to handle variable image sizes naturally without fixed patching. It uses a customized SigLIP-based visual encoder (approx. 600M params) with a C-Abstractor for feature compression. The backbone is Qwen2.5-7B, consisting of 28 layers and 28 attention heads.

\subsection{Hallucination Detection Baselines}
\label{baseline}

\paragraph{AvgProb}
Given a generated sentence (or sequence) indexed by $i$ with $J_i$ tokens, let $p_{ij}$ denote the model-assigned conditional probability of the \emph{actually generated} token at position $j$.
AvgProb quantifies sentence-level uncertainty by the mean negative log-probability over all positions:
\[
\mathrm{AvgProb}(i) \;=\; -\frac{1}{J_i}\sum_{j=1}^{J_i}\log p_{ij}.
\]
A larger $\mathrm{AvgProb}(i)$ indicates that the model tends to assign lower likelihood to the produced tokens, reflecting higher uncertainty for the whole sentence.

\paragraph{AvgEnt}
AvgEnt computes uncertainty using the full predictive distribution at each position.
Let $\mathbf{p}_{ij}(\cdot)$ be the predicted distribution over the vocabulary $\mathcal{V}$ at position $j$ in sentence $i$, and define the token-level predictive entropy as
\[
H_{ij} \;=\; -\sum_{v\in\mathcal{V}} \mathbf{p}_{ij}(v)\,\log \mathbf{p}_{ij}(v).
\]
We then aggregate token entropies into a sentence-level score via averaging:
\[
\mathrm{AvgEnt}(i) \;=\; \frac{1}{J_i}\sum_{j=1}^{J_i} H_{ij}.
\]
Higher $\mathrm{AvgEnt}(i)$ suggests more diffuse (less confident) predictive distributions across tokens, hence greater sentence-level uncertainty.

\paragraph{RepProbing}
RepProbing includes a lightweight classifier trained on the VLM decoder's last-layer hidden states to estimate hallucination risk.
Let $z_t^{L}\in\mathbb{R}^{d}$ be the hidden state at token position $t$ from the top decoder layer $L$.
The probe outputs a hallucination score (or probability) as
\begin{equation}
\hat{y}^{\,h}_t \;=\; f_{\theta}\!\left(z_t^{L}\right),
\label{eq:last_layer_probe}
\end{equation}
where $f_{\theta}$ is typically a linear head or a shallow MLP.

\subsection{Hallucination Mitigation Baselines}

\paragraph{BeamSearch}
Beam search is a deterministic decoding strategy that approximates the most likely output sequence by maintaining the top-$B$ partial hypotheses (``beams'') at each step.
Starting from the prompt, it repeatedly expands each beam with candidate next tokens and keeps only the $B$ sequences with the highest cumulative log-probability (often with length normalization), continuing until an end-of-sequence token is produced.

\paragraph{PAI}
PAI \cite{liu2024paying} is a training-free method that mitigates text inertia in LVLMs—when the LLM dominates so outputs rely more on text context than visual evidence. It boosts attention to image tokens and subtracts text-only logits from multimodal logits to suppress language-only bias, encouraging stronger visual grounding and reducing hallucinations.

\paragraph{VCD}
VCD (Visual Contrastive Decoding) \cite{leng2024mitigating} is a simple, training-free decoding method that contrasts the output distributions produced from an original image and a distorted version of the same image. By using this contrast to suppress statistical biases and unimodal language priors, it encourages stronger visual grounding, substantially reducing object hallucinations across LVLM families while also performing well on general LVLM benchmarks.

\section{Implementation Details}
\subsection{Hallucination Detection}
\label{setting}
In the hallucination detection experiments, for the discriminative benchmarks POPE and AMBER, we follow the work of \cite{Li_2024} to extract images, questions, and ground truths (GT) from the original datasets. For each sample, we construct responses that either contain or do not contain hallucinations; specifically, for samples where the GT is ``Yes'', we generate ``Yes'' (containing hallucination) and ``No'' (free from hallucination) responses.

For the POPE benchmark, we construct training and validation splits across its three subsets (popular, random, and adversarial) and report the average metrics over these subsets. For the AMBER benchmark, we conduct experiments using a curated subset of 5,000 samples. We manually partition the datasets to ensure that different samples associated with the same image do not overlap between the training and validation sets.

For the M-HalDetect benchmark, we further divide the official validation set into training and validation subsets using an $80:20$ ratio and report span-based hallucination detection results. Regarding the COCO-Caption task, we employ the LLaVA-v1.5-7B model to generate responses for images from the COCO 2014 Val set. We annotate hallucinated objects in the responses using the official COCO 2014 Val annotations and report sentence-based hallucination detection results.

\subsection{Model Architecture}
Regarding the VIB-Probe encoder, we utilize a 3-layer MLP network with dimensions $(1024, 512, 256)$ to reduce the dimensionality of the original attention output feature vectors, followed by processing with two residual blocks. For the decoder, we employ a simple single linear layer. Throughout the network, we apply the GELU activation function and LayerNorm.

\subsection{Hallucination Mitigation}
We evaluate object hallucinations in VLM's generation with the CHAIR (Captioning Hallucination Assessment with Image Relevance) metrics, which compare model-generated captions against ground-truth object annotations to quantify objects mentioned in text but not present in the image. Specifically, CHAIR$_i$ reports the proportion of hallucinated object mentions among all generated object mentions, while CHAIR$_s$ reports the percentage of captions that contain at least one hallucinated object.
\begin{equation}
\small
\mathrm{CHAIR}_{i} = 
\frac{|\{\text{hallucinated objects}\}|}
     {|\{\text{all objects mentioned}\}|}
\end{equation}
\begin{equation}
\small
\mathrm{CHAIR}_{s} = 
\frac{|\{\text{sentences with hallucinated objects}\}|}
     {|\{\text{all sentences}\}|}
\end{equation}

In the hallucination mitigation experiments, we intervene on the attention heads that rank in the top $5\%$ of head importance scores. The threshold for triggering this intervention is determined based on the average logit values from the training set used in the hallucination detection experiments. We set the suppression strength $\lambda$ to $0.001$.


\end{document}